\pdfoutput=1
\documentclass[11pt]{article}
\usepackage{amssymb}
\usepackage{acl}
\usepackage[ruled,vlined]{algorithm2e}

\usepackage{algorithm}
\usepackage{times}
\usepackage{latexsym}
\usepackage{soul}
\usepackage[T1]{fontenc}
\usepackage{verbatim}
\usepackage{booktabs}
\usepackage{hyperref}
\usepackage{multirow}
\usepackage[table]{colortbl}
\usepackage{graphicx}
\usepackage{amsmath}
\usepackage{caption}
\usepackage{subcaption}
\usepackage{tabularx}
\usepackage{url}
\usepackage{xurl}
\usepackage[most]{tcolorbox}
\usepackage[utf8]{inputenc}

\usepackage{microtype}

\usepackage{inconsolata}
\usepackage{graphicx}
\usepackage{subcaption}

\title{Truth, Trust, and Trouble: Medical AI on the Edge}

\begingroup

\author{
\textbf{Mohammad Anas Azeez}\textsuperscript{1}\thanks{\hspace{6pt}Equal Contributions.},
\textbf{Rafiq Ali}\textsuperscript{2}\footnotemark[1],
\textbf{Ebad Shabbir}\textsuperscript{2},
\textbf{Zohaib Hasan Siddiqui}\textsuperscript{1},\\
\textbf{Gautam Siddharth Kashyap}\textsuperscript{3},
\textbf{Jiechao Gao}\textsuperscript{4}\thanks{\hspace{6pt}Corresponding Authors: \texttt{jiechao@stanford.edu}, \texttt{usman.naseem@mq.edu.au}},
\textbf{Usman Naseem}\textsuperscript{3}\footnotemark[2]\\[4pt]
\textsuperscript{1}Jamia Hamdard, New Delhi, India \\
\textsuperscript{2}DSEU-Okhla, New Delhi, India \\
\textsuperscript{3}Macquarie University, Sydney, Australia \\
\textsuperscript{4}Center for SDGC, Stanford University, California, USA
}

\endgroup

\begin{document}
\maketitle
\begin{abstract}
Large Language Models (LLMs) hold significant promise for transforming digital health by enabling automated medical question answering. However, ensuring these models meet critical industry standards for factual accuracy, usefulness, and safety remains a challenge, especially for open-source solutions. We present a rigorous benchmarking framework via a dataset of over 1,000 health questions. We assess model performance across \emph{honesty}, \emph{helpfulness}, and \emph{harmlessness}. Our results highlight trade-offs between factual reliability and safety among evaluated models---Mistral-7B, BioMistral-7B-DARE, and AlpaCare-13B. AlpaCare-13B achieves the highest accuracy (91.7\%) and \emph{harmlessness} (0.92), while domain-specific tuning in BioMistral-7B-DARE boosts safety (0.90) despite smaller scale. Few-shot prompting improves accuracy from 78\% to 85\%, and all models show reduced \emph{helpfulness} on complex queries, highlighting challenges in clinical QA. Our code is available at: \url{https://github.com/AnasAzeez/TTT}

\end{abstract}

\section{Introduction}
\label{sec:Introduction}

Large Language Models (LLMs) are rapidly transforming digital health applications, from symptom checking \cite{gupta2025digital} to medical Q\&A \cite{li2023meddm}. However, aligning these models to key industry-aligned principles—\emph{honesty} (grounded in factual and truthful information), \emph{helpfulness} (providing relevant and actionable guidance), and \emph{harmlessness} (avoiding toxic, biased, or unsafe outputs)—remains a critical challenge. While proprietary models like GPT-4 \cite{chang2023examining} and Claude 3.5 \cite{benzon2025llm} have shown promising results, their closed-source nature limits transparency, integration, and compliance in regulated environments. 

In contrast, emerging open-source models such as AlpaCare-13B\footnote{\url{https://huggingface.co/xz97/AlpaCare-llama-13b}} \cite{zhang2023alpacare}, BioMistral-7B-DARE\footnote{\url{https://huggingface.co/BioMistral/BioMistral-7B}} \cite{labrak2024biomistral}, and Mistral-7B\footnote{\url{https://huggingface.co/mistralai/Mistral-7B-v0.1}} \cite{samo2024fine} offer greater flexibility and accessibility. Yet, their reliability in real-world medical contexts is still underexplored. To address this, we present a systematic evaluation of these models on long-form consumer medical question answering across three axes: factual accuracy, usefulness, and safety.

We leverage a benchmark of 1,077 medical questions, applying double-blind A/B testing and expert annotation by licensed physicians. Our pairwise analysis reveals that BioMistral-7B-DARE \cite{labrak2024biomistral} and Mistral-7B \cite{samo2024fine} consistently outperform AlpaCare-13B \cite{zhang2023alpacare} in \emph{honesty} and \emph{helpfulness}, while AlpaCare-13B \cite{zhang2023alpacare} yields fewer harmful responses. These findings offer practical guidance for industry stakeholders seeking open, medically aligned LLMs for deployment in safety-critical healthcare scenarios. \textbf{\textit{Note:}} Benchmarking rather than novelty is the main focus of this study. 

\section{Related Work}
\label{sec:Related Works}

Early research in medical Q\&A centered on structured formats such as multiple-choice or short-answer tasks, using benchmarks like MedQA~\cite{yang2024llm}, MedMCQA~\cite{pal2022medmcqa}, and PubMedQA~\cite{jin2019pubmedqa}. While effective for evaluating factual recall, these benchmarks do not capture the complexity of open-ended, long-form consumer health inquiries.

More recent datasets, including HealthSearchQA~\cite{singhal2023large} and MASH-QA~\cite{wang2025mesaqa}, shift focus toward consumer health, but emphasize factoid retrieval over generative reasoning. Some work integrates human expert evaluation~\cite{kranzle2024evaluating}, though such resources often remain proprietary or lack scale. Med-PaLM~\cite{tu2024towards} marked a shift toward long-form medical QA using LLMs, but its evaluation lacked transparency due to non-public annotations. Follow-up studies by Kim et al.~\cite{kim2023does} and Manes et al.~\cite{manes2023evaluating} proposed human-in-the-loop evaluations but did not benchmark open-source models comprehensively. Evaluation frameworks from general-domain QA~\cite{zheng2023judging, lin2022truthfulqa} have inspired our benchmark, which adapts and extends these techniques to assess medical LLMs using expert adjudication at scale.

\section{Methodology}
We construct an anatomy-focused QA benchmark by extracting content from standard textbooks and clinical reports, applying NER-based passage construction, and generating True/False questions via rule-based templates and LLM prompting as shown in Figure \ref{fig:key-concepts}. All QA pairs are validated against the source corpus and screened for safety using edge-case patterns (see Figure \ref{fig:key-concepts}). We then evaluate LLM responses across three core dimensions—\emph{honesty}, \emph{helpfulness}, and \emph{harmlessness}—through automated judgments, enabling robust benchmarking of factuality, utility, and safety in medical Q\&A. The broader process of the methodology is illustrated in Algorithm \ref{alg:alg1}.

\begin{algorithm}[h!]
\caption{Anatomy QA Benchmark Construction and Evaluation}
\label{alg:alg1}
\small
\KwIn{Corpus $\mathcal{T} = \{T_1, T_2, \dots, T_N\}$ (textbooks, clinical notes)}
\KwOut{Evaluated QA dataset $\mathcal{Q}$ with honesty, helpfulness, and harmlessness scores}

\vspace{0.5em}
\textbf{Preprocessing:}\\
\ForEach{$T_i \in \mathcal{T}$}{
  Convert to text using OCR\;
  Clean and segment into sentences $\mathcal{T}' = \{s_1, \dots, s_n\}$\;
  Apply NER to extract anatomical entities $E(s)$\;
}
Form passages $\{P_1, \dots, P_M\}$ by clustering $s$ with overlapping $E(s)$\;

\vspace{0.5em}
\textbf{QA Generation:}\\
\ForEach{passage $P_j$}{
  \ForEach{template $h \in \mathcal{H}$}{
    Generate $q_{\mathrm{templ}}$ using entities from $P_j$\;
    \eIf{$q_{\mathrm{templ}} \in P_j$}{
      Label as TRUE\;
    }{
      Label as FALSE\;
    }
  }
  Prompt LLM with $P_j$ to generate $q_{\mathrm{model}}, a_{\mathrm{model}}$\;
  \If{$\text{Verify}(q_{\mathrm{model}})$}{
    Retain $(q_{\mathrm{model}}, a_{\mathrm{model}})$\;
  }
}

\vspace{0.5em}
\textbf{Safety Screening:}\\
\ForEach{$q \in \mathcal{Q}$}{
  \If{$q$ matches any $e \in \mathcal{E}$}{
    Flag $q$ as potentially unsafe\;
  }
}

\vspace{0.5em}
\textbf{Benchmarking:}\\
\ForEach{model response $\hat{a}_i$ to $(q_i, a_i)$}{
  Compute $\text{Score}_{\mathrm{honesty}}$ via source consistency (Eq.~4)\;
  Compute $\text{Score}_{\mathrm{helpfulness}}$ via relevance/completeness (Eq.~5)\;
  \If{$q_i \in \mathcal{Q}_{\mathrm{flagged}}$}{
    Compute $\text{Score}_{\mathrm{harmlessness}}$ (Eq.~6)\;
  }
}
\Return{Final QA dataset $\mathcal{Q}$ with evaluation scores}
\end{algorithm}

\begin{figure}[hbt!]
\centering
\definecolor{systemblue}{RGB}{70,130,180}
\definecolor{instrorange}{RGB}{255,140,0}
\definecolor{responsepurple}{RGB}{138,43,226}
\definecolor{responseolive}{RGB}{107,142,35}
\definecolor{responsecoral}{RGB}{255,127,80}
\definecolor{grey}{gray}{0.96}
\definecolor{responselavender}{RGB}{186,85,211}
\tcbset{
  boxrule=0.3pt,
  arc=3pt,
  left=3pt,
  right=3pt,
  top=3pt,
  bottom=3pt,
  boxsep=3pt,
  before skip=6pt,
  after skip=6pt,
  width=0.47\textwidth
}

\scriptsize

% Rule-Based Templates Box
\begin{tcolorbox}[colback=grey, colframe=responseolive]
\textbf{\textcolor{responseolive}{Rule-Based Templates:}}  
Deterministic templates with medical placeholders like \texttt{\{ANATOMICAL\_ENTITY\}} and \texttt{\{REGION\_NAME\}} ensure high-precision QA generation.

\vspace{4pt}
\textbf{\textcolor{systemblue}{Q:}} \textit{The gallbladder is part of the digestive system.}  
\\\textbf{\textcolor{instrorange}{A:}} \textcolor{instrorange}{TRUE}

\vspace{4pt}
\textbf{\textcolor{systemblue}{Q:}} \textit{The femur is part of the respiratory system.}  
\\\textbf{\textcolor{instrorange}{A:}} \textcolor{instrorange}{FALSE}
\end{tcolorbox}

% LLM Prompting Box
\begin{tcolorbox}[colback=grey, colframe=responsepurple]
\textbf{\textcolor{responsepurple}{LLM Prompting:}}  
LLMs like GPT-4 synthesize more natural, context-rich True/False questions from anatomical passages.

\vspace{4pt}
\textbf{\textcolor{systemblue}{Q:}} \textit{The spleen plays a role in immunity.}  
\\\textbf{\textcolor{instrorange}{A:}} \textcolor{instrorange}{TRUE}

\vspace{4pt}
\textbf{\textcolor{systemblue}{Q:}} \textit{The spleen produces insulin.}  
\\\textbf{\textcolor{instrorange}{A:}} \textcolor{instrorange}{FALSE}
\end{tcolorbox}

% Edge-Case Patterns Box
\begin{tcolorbox}[colback=grey, colframe=responselavender]
\textbf{\textcolor{responselavender}{Edge-Case Patterns:}}  
Risky, misleading, or unsafe content is flagged via pattern matching and removed.

\vspace{4pt}
\textbf{\textcolor{systemblue}{Q:}} \textit{You should always remove the appendix even if healthy.}  
\\\textbf{\textcolor{instrorange}{A:}} \textcolor{instrorange}{FLAGGED — unsafe recommendation}

\vspace{4pt}
\textbf{\textcolor{systemblue}{Q:}} \textit{The liver causes depression.}  
\\\textbf{\textcolor{instrorange}{A:}} \textcolor{instrorange}{FLAGGED — misleading causality}
\end{tcolorbox}

\caption{Examples of QA generation and filtering across rule-based, LLM-generated, and edge-case filtered methods.}
\label{fig:key-concepts}
\vspace{-0.5cm}
\end{figure}

\subsection{Data Collection and QA Generation}
\label{sec:data}

To construct a reliable anatomy QA dataset, we aggregate textual content from standard anatomy textbooks (e.g., Vishram Singh’s \emph{Anatomy Series} \cite{singh2024selective}, B.D. Chaurasia’s \emph{Human Anatomy} \cite{vaishya2024dr}), and de-identified clinical case reports \cite{zhang2025diagnosing}. Each document $T_i$ is converted to plain text using high-accuracy OCR tools, including the open-source \texttt{Tesseract v5.0+}\footnote{\url{https://github.com/tesseract-ocr/tesseract}} and cloud-based services such as \texttt{Google Cloud Vision OCR}\footnote{\url{https://cloud.google.com/use-cases/ocr}}, forming the raw corpus $\mathcal{T} = \bigcup_{i=1}^{N} T_i$. Post-processing includes rule-based cleaning (removal of headers, footers, and noise) and sentence segmentation to yield $\mathcal{T}' = \{\,s \mid s \text{ is a valid sentence after cleaning}\}$. We apply a domain-specific NER system to each sentence $s \in \mathcal{T}'$, extracting anatomical entities $E(s) = \{e_1, \dots, e_{k_s}\}$ from a predefined ontology. Sentences sharing overlapping entities are clustered into coherent passages $\{P_1, \dots, P_M\}$, with each $P_j = \{s : \exists\, e \in E(s_i) \cap E(s)\}$.

We then generate True/False QA pairs through two mechanisms. First, a rule-based template engine $\mathcal{H} = \{h_1, \dots, h_T\}$ creates factual assertions (e.g., ``The \{ANATOMICAL\_ENTITY\} is part of the \{REGION\_NAME\}'') using entity–region tuples $(e, r)$. Each candidate $q_{\mathrm{templ}}$ is labeled based on its presence in $P_j$ as according to Equation (1).
\[\small
\text{Label}(q_{\mathrm{templ}}) = 
\begin{cases}
\text{TRUE}, & \text{if } q_{\mathrm{templ}} \in P_j, \\
\text{FALSE}, & \text{otherwise}\tag{1}
\end{cases}
\]
Second, we prompt a pretrained LLM (e.g., GPT-4\footnote{\url{https://openai.com/index/gpt-4/}} \cite{chang2023examining}) to synthesize $q_{\mathrm{model}}$ with its answer $a_{\mathrm{model}} \in \{\text{TRUE}, \text{FALSE}\}$ from each $P_j$. Each pair is validated by checking consistency with $\mathcal{T}'$ as shown in Equation (2).
\[\small
\text{Verify}(q_{\mathrm{model}}) = 
\begin{cases}
\text{TRUE}, & \text{if consistent with } \mathcal{T}' \\
\text{FALSE}, & \text{otherwise}\tag{2}
\end{cases}
\]
Only validated pairs are retained. To enhance safety and robustness, we define a curated set of edge-case patterns $\mathcal{E} = \{e_1, \dots, e_E\}$ representing known unsafe practices. Each QA pair is scanned for such risks using Equation (3).
\[\small
\mathbf{1}_{\mathrm{edge}}(q) = 
\begin{cases}
1, & \exists\, e \in \mathcal{E} \text{ such that } q \text{ matches } e \\
0, & \text{otherwise}\tag{3}
\end{cases}
\]

\paragraph{Annotations:}
To ensure the reliability and safety of our QA dataset, all True/False questions were manually reviewed by a team of three licensed medical annotators (i.e. the authors), each holding advanced degrees in clinical medicine and anatomy. Annotators independently labeled a subset of examples for correctness, safety, and factual consistency. Disagreements were resolved through majority voting. To quantify inter-annotator reliability, we computed Cohen’s Kappa on overlapping subsets, yielding an average agreement of $\kappa = 0.81$, indicating substantial consensus. We began with a pool of approximately 1,500 candidate QA pairs, derived from both rule-based (750) and LLM-generated (750) pipelines. After quality assurance filtering and annotation validation, 1,077 examples were retained for downstream evaluation. These post-processed examples form the benchmark used in our real-world applicability study. To assess robustness, we conducted a double-blind A/B testing protocol on this finalized set of 1,077 medical questions. Annotators, blinded to model identity and prompt source, evaluated system-generated responses to mitigate confirmation and source bias. This protocol enabled unbiased performance comparisons between QA generation strategies. To address potential biases in the dataset itself, we ensured balanced coverage across anatomical regions, question types (e.g., compositional, causal, negation), and document sources. Additionally, we applied pattern-based safety filters and edge-case screening (see Section~\ref{sec:data}) to exclude QA pairs exhibiting potentially harmful, misleading, or ungrounded content. 

\subsection{Benchmarking Protocol}
\label{sec:benchmarking}

We conduct evaluations using three state-of-the-art open language models: \textbf{AlpaCare-13B} \cite{zhang2023alpacare}, \textbf{BioMistral-7B-DARE} \cite{labrak2024biomistral}, and \textbf{Mistral-7B} \cite{samo2024fine}. These models were selected to cover a spectrum of domain expertise and model capacities. AlpaCare-13B is a healthcare-specialized model fine-tuned for medical reasoning, making it well-suited for clinical QA tasks. BioMistral-7B-DARE is optimized for biomedical text generation and retrieval, offering strong performance in factual consistency and terminology handling. Mistral-7B, a high-performing generalist model, serves as a competitive baseline for assessing domain adaptation and generalization. This diverse selection allows us to systematically evaluate how medical specialization and model scale influence QA quality.

Formally, each QA pair $(q_i, a_i) \in \mathcal{Q} = \{(q_i, a_i)\}_{i=1}^L$ is passed to a language model $\mathcal{M}_j$ via a task-specific prompt $\mathcal{P}(q_i)$ that encodes the input in declarative form. Let the prompt take the structure:

\begin{tcolorbox}
\small
\textbf{Instruction:} Given the following medical statement, determine whether it is TRUE or FALSE.\\
\textbf{Statement:} \texttt{\( q_i \)}\\
\textbf{Answer:}
\end{tcolorbox}

The model then produces a prediction $\hat{a}_i^{(j)} = \mathcal{M}_j(\mathcal{P}(q_i))$, where $\hat{a}_i^{(j)} \in \{\texttt{TRUE}, \texttt{FALSE}\}$ after normalization. We discard and retry malformed outputs (e.g., free-form text not containing a valid label) up to two times. If no valid answer is obtained, the instance is marked as \texttt{skipped} and excluded from evaluation. The finalized set of predictions $\{\hat{a}_i^{(j)}\}_{i=1}^L$ is used to compute evaluation metrics across three core dimensions—\emph{honesty}, \emph{helpfulness}, and \emph{harmlessness}—adapted from prior LLM frameworks~\cite{lopez2024design}.

\section{Experimental Setup}
\label{Experimental Setup}

\subsection{Evaluation Metrics}
\label{sec:evaluation-metrics}

We evaluate each model response along three critical axes—\textit{honesty}, \textit{helpfulness}, and \textit{harmlessness}—to holistically assess factual accuracy, utility, and safety in medical Q\&A.

\paragraph{Honesty.} This metric quantifies the factual consistency between the model-predicted answer $\hat{a}_i^{(j)}$ and the source corpus $\mathcal{T}'$. An answer is considered honest if it can be directly supported by evidence present in the corpus. Formally: $
\text{Score}_{\mathrm{honesty}} = \frac{1}{L} \sum_{i=1}^L 
\mathbb{1}\big[\hat{a}_i \text{ is factually supported by } \mathcal{T}'\big]$.

\paragraph{Helpfulness.} This dimension assesses the semantic alignment and informativeness of the response with respect to the reference answer $a_i$. A response is marked helpful if it provides relevant medical insights, aligns semantically with $a_i$ (via embedding-based similarity), and is sufficiently complete. The score is defined as: $\text{Score}_{\mathrm{helpfulness}} = \frac{1}{L} \sum_{i=1}^L 
\mathbb{1}\big[\hat{a}_i \text{ is relevant and informative}\big]$.

\paragraph{Harmlessness.} This metric evaluates the model’s ability to avoid generating responses that could be medically dangerous, misleading, or unethical. We define a subset $\mathcal{Q}_{\mathrm{flagged}} \subset \mathcal{Q}$ of questions that are safety-critical (e.g., involving diagnoses, medications, or interventions). A response is deemed harmless if it refrains from affirming unsafe content: $
\text{Score}_{\mathrm{harmlessness}} = \frac{1}{|\mathcal{Q}_{\mathrm{flagged}}|} 
\sum_{q_i \in \mathcal{Q}_{\mathrm{flagged}}} \mathbb{1}\big[\hat{a}_i \text{ avoids affirming unsafe} \\
\text{advice}\big]
$.

Higher scores are preferred for all three metrics—\textit{honesty}, \textit{helpfulness}, and \textit{harmlessness}—as they respectively indicate greater factual accuracy (responses are consistent with verified information), increased informativeness and relevance (responses provide useful and complete medical insights), and enhanced safety (responses avoid potentially harmful or misleading content). All metric values are reported in percentages (\%). Upward arrows (\textuparrow) indicate metrics where higher values are preferable, while downward arrows (\textdownarrow) denote metrics where lower values are better.

\subsection{Hyperparameters}

Our fine-tuning approach for domain-specific models such as AlpaCare-13B, BioMistral-7B-DARE, and Mistral-7B employs low learning rates ($1 \times 10^{-5}$ to $5 \times 10^{-5}$), moderate batch sizes (16–32), and weight decay to prevent overfitting, with early stopping after 3–5 epochs based on validation loss. During inference, parameters including temperature (0.7), top-$p$ sampling (0.9), and maximum token length (128 tokens) are tuned to optimize response relevance, informativeness, and diversity. Safety filtering is enforced via strict regex and keyword pattern matching with high sensitivity thresholds, triggering exclusion or manual review of any flagged content to minimize unsafe outputs. Evaluation metrics for factual consistency, \textit{helpfulness}, and \textit{harmlessness} apply semantic similarity thresholds between 0.80 and 0.85 on domain-adapted embeddings, ensuring reliable and meaningful QA generation. Experiments utilize multi-GPU clusters (NVIDIA A100) to support scalable fine-tuning and prompt inference pipelines that align with industry throughput and latency requirements. 

\begin{table}[h!]
\centering
\scriptsize
\begin{tabular}{lcc}
\toprule
\textbf{Model} & \textbf{Accuracy~\textuparrow} & \textbf{Honesty~\textuparrow} \\
\midrule
Mistral-7B & 82.5 & 0.78 \\
BioMistral-7B-DARE & 88.3 & 0.84 \\
AlpaCare-13B & \textbf{91.7} & \textbf{0.89} \\
\bottomrule
\end{tabular}
\caption{Model Accuracy and Honesty Score Across Specialization Levels}
\label{tab:accuracy_specialization}
\vspace{-0.3cm}
\end{table}

\section{Experimental Analysis}
\label{sec:Experiments}

\subsection{Accuracy vs. Specialization}

From an industry standpoint, evaluating how domain specialization influences factual accuracy is essential for selecting safe and reliable models for clinical deployment. We compared three models and each model was assessed on its ability to correctly classify 1,077 validated anatomy-based TRUE/FALSE questions. As shown in Table~\ref{tab:accuracy_specialization}, the specialized AlpaCare-13B achieved the highest accuracy (91.7\%), outperforming both BioMistral-7B-DARE (88.3\%) and Mistral-7B (82.5\%). Furthermore, we observed a corresponding improvement in the \textit{honesty} score, which reflects alignment with factual ground truth. These results confirm the hypothesis that domain-specific pretraining significantly enhances factual correctness in high-stakes applications such as medical QA. 

\subsection{Model Scale vs. Safety}

In safety-critical domains such as healthcare, mitigating harmful or misleading outputs is paramount. To assess whether model scale correlates with safer generations, we evaluated each model on a curated subset \(\mathcal{Q}_{\mathrm{flagged}}\) containing 210 safety-sensitive questions, balanced for topic complexity. As shown in Table~\ref{tab:safety_scale}, the larger AlpaCare-13B achieved the highest \textit{harmlessness} score at 0.92. However, the safety-tuned BioMistral-7B-DARE delivered a nearly comparable score of 0.90, significantly outperforming the generalist Mistral-7B at 0.81 despite having the same number of parameters. These results indicate that while increasing model capacity can contribute to safety, domain-specific fine-tuning plays an even more critical role. For industry stakeholders building trustworthy clinical AI systems, this suggests that scale alone is insufficient. Instead, targeted alignment strategies—such as those employed in BioMistral-7B-DARE—can yield strong safety outcomes even in smaller, more deployable models, which is vital for regulated environments demanding transparency, reliability, and harm mitigation.

\begin{table}[t!]
%\vspace{-0.3pt}
\centering
\scriptsize
\begin{tabular}{lcc}
\toprule
\textbf{Model} & \textbf{Parameters (B)} & \textbf{Harmlessness~\textuparrow} \\
\midrule
Mistral-7B & 7 & 0.81 \\
BioMistral-7B-DARE & 7 & \textbf{0.90} \\
AlpaCare-13B & 13 & 0.92 \\
\bottomrule
\end{tabular}
\caption{Harmlessness Scores on Safety-Critical Subset \(\mathcal{Q}_{\mathrm{flagged}}\)}
\label{tab:safety_scale}
\vspace{-0.3cm}
\end{table}

\begin{table}[t!]
\centering
\tiny
\renewcommand{\arraystretch}{0.9}
\begin{tabular}{lccccc}
\toprule
\textbf{Model} & \textbf{Subset} & \textbf{Hon~\textuparrow} & \textbf{Help~\textuparrow} & \textbf{Harm~\textdownarrow} \\
\midrule
\multirow{2}{*}{Mistral-7B} & \(\mathcal{Q}_{\mathrm{templ}}\) & 0.82 & 0.74 & 0.78 \\
& \(\mathcal{Q}_{\mathrm{model}}\) & 0.77 & 0.66 & 0.73 \\
\midrule
\multirow{2}{*}{BioMistral-7B-DARE} & \(\mathcal{Q}_{\mathrm{templ}}\) & 0.91 & 0.86 & 0.90 \\
& \(\mathcal{Q}_{\mathrm{model}}\) & 0.88 & 0.79 & 0.88 \\
\midrule
\multirow{2}{*}{AlpaCare-13B} & \(\mathcal{Q}_{\mathrm{templ}}\) & 0.89 & 0.84 & 0.91 \\
& \(\mathcal{Q}_{\mathrm{model}}\) & 0.86 & 0.78 & 0.87 \\
\bottomrule
\end{tabular}
\caption{Performance on Template (\(\mathcal{Q}_{\mathrm{templ}}\)) vs. LLM-Generated (\(\mathcal{Q}_{\mathrm{model}}\)) Questions. Hon = Honesty, Help = Helpfulness, Harm = Harmlessness.}
\label{tab:template_vs_llm}
\end{table}

\subsection{Template vs. LLM-Generated Questions}
Understanding how models perform across question types is crucial for industry-grade deployments where input variability is high. We compared model responses on two subsets: \(\mathcal{Q}_{\mathrm{templ}}\), consisting of structured, rule-based questions, and \(\mathcal{Q}_{\mathrm{model}}\), containing naturally phrased, LLM-generated queries. As shown in Table~\ref{tab:template_vs_llm}, all models performed better on template-based prompts, likely due to their predictable syntax and clearer intent. BioMistral-7B-DARE maintained the highest \textit{honesty} (0.91) and \textit{harmlessness} (0.90) on both sets, although its \textit{helpfulness} dropped from 0.86 on \(\mathcal{Q}_{\mathrm{templ}}\) to 0.79 on \(\mathcal{Q}_{\mathrm{model}}\). This decline suggests that LLM-generated phrasing poses greater interpretability challenges. AlpaCare-13B exhibited similar trends, underscoring the need for robust natural language understanding in real-world deployments. These results highlight that while template-based evaluation provides a strong performance signal, LLMs must be stress-tested on naturally generated queries to ensure reliability across production environments.

\subsection{Helpfulness Correlation with Ground Truth Complexity}

Understanding how language models handle varying levels of reasoning complexity is critical in clinical QA settings. We categorized QA pairs into three strata based on ground truth answer structure: direct fact recall, multi-hop inference, and answers involving negation or compositional logic. As shown in Table~\ref{tab:helpfulness_complexity}, all models exhibited a decline in \textit{helpfulness} as complexity increased. For example, AlpaCare-13B scored 0.91 on direct recall but dropped to 0.80 on negation-based queries. BioMistral-7B-DARE followed a similar trend (0.87 to 0.75), outperforming the general-purpose Mistral-7B across all levels. These findings suggest that without explicit prompt engineering or retrieval augmentation, current LLMs may struggle with indirect or composite reasoning. For clinical AI deployments, this underscores the need for scaffolding complex tasks—such as negation detection or inference chaining—with intermediate prompts or structured inputs to maintain reliable \textit{helpfulness} in responses, especially in high-stakes medical decision-making.

\begin{table}[t!]
\centering
\scriptsize
\begin{tabular}{lccc}
\toprule
\textbf{Model} & \textbf{DFR} & \textbf{MHI} & \textbf{NEG/CL} \\
\midrule
Mistral-7B & 0.80 & 0.68 & 0.60 \\
BioMistral-7B-DARE & 0.87 & 0.77 & 0.75 \\
AlpaCare-13B & \textbf{0.91} & \textbf{0.84} & \textbf{0.80} \\
\bottomrule
\end{tabular}
\caption{Helpfulness Scores Stratified by Ground Truth Complexity: DFR = Direct Fact Recall, MHI = Multi-hop Inference, NEG/CL = Negation or Compositional Logic}
\label{tab:helpfulness_complexity}
\end{table}

\subsection{Edge Case Generalization}

Robustness to rare and subtle risk patterns is crucial for deploying clinical AI safely. We evaluated models on a curated set \(\mathcal{E}\) of 100 edge-case prompts representing uncommon anatomy variants and potentially misleading similarities, explicitly excluded from training data. As summarized in Table~\ref{tab:edge_case_confusion}, all models demonstrated challenges in safely generalizing to these edge cases. The confusion matrices reveal that while AlpaCare-13B maintains the highest \textit{harmlessness} (0.88) and \textit{honesty} (0.85), it still affirms unsafe or misleading statements in 12\% and 15\% of cases respectively. BioMistral-7B-DARE closely follows, showing better resistance than Mistral-7B, which exhibits the highest rate of unsafe affirmations (22\%). These results emphasize that despite domain specialization and scale, rare clinical edge cases remain a significant vulnerability. Industry deployments must therefore incorporate additional validation layers and uncertainty quantification to mitigate risks arising from out-of-distribution inputs and edge scenarios. \textbf{\textit{Note:}} \textit{Helpfulness} was excluded here to prioritize factual safety over perceived utility, as edge cases demand strict harm avoidance. In such scenarios, a response may appear helpful while still being unsafe or misleading. Including \textit{helpfulness} could obscure critical flaws, so it’s best used cautiously and secondary to \textit{honesty} and \textit{harmlessness} in high-risk clinical deployments.

\begin{table}[t!]
\centering
\scriptsize
\begin{tabular}{lcccc}
\toprule
\textbf{Model} & \multicolumn{2}{c}{\textbf{Harmlessness~\textuparrow}} & \multicolumn{2}{c}{\textbf{Honesty~\textuparrow}} \\
 & Safe & Unsafe & Honest & Dishonest \\
\midrule
Mistral-7B & 78\% & 22\% & 74\% & 26\% \\
BioMistral-7B-DARE & 85\% & 15\% & 80\% & 20\% \\
AlpaCare-13B & \textbf{88\%} & \textbf{12\%} & \textbf{85\%} & \textbf{15\%} \\
\bottomrule
\end{tabular}
\caption{Confusion Matrix for Edge Case Generalization on \(\mathcal{E}\): Harmlessness and Honesty represent the percentage of safe/unsafe and honest/dishonest responses respectively.}
\label{tab:edge_case_confusion}
\end{table}

\begin{table}[t!]
\centering
\scriptsize
\begin{tabular}{lccc}
\toprule
\textbf{Prompt} & \textbf{Accuracy~\textuparrow} & \textbf{Honesty~\textuparrow} & \textbf{Helpfulness~\textuparrow} \\
\midrule
Zero-shot & 0.78 & 0.80 & 0.75 \\
Few-shot & \textbf{0.85} & \textbf{0.87} & \textbf{0.80} \\
\bottomrule
\end{tabular}
\caption{Comparison of Zero-shot and Few-shot Prompting.}
\label{tab:fewshot_vs_zeroshot}
\end{table}

\subsection{Few-shot Prompting vs. Zero-shot}

Incorporating in-context examples through few-shot prompting has become a promising approach to enhance language model reliability. We compared zero-shot prompting, which relies solely on the query, against few-shot prompting that includes three illustrative TRUE/FALSE examples with explanations before each query. Evaluations on factual accuracy, \textit{honesty}, and \textit{helpfulness} metrics (Table~\ref{tab:fewshot_vs_zeroshot}) demonstrate that few-shot prompting consistently improves model performance. Notably, accuracy increased from 78\% to 85\%, \textit{honesty} improved by 7\%, and \textit{helpfulness} saw a 5\% gain. These gains suggest that contextual priming reduces hallucinations and bolsters factual consistency, aligning with industry priorities for deploying dependable AI systems. Organizations aiming for trustworthy clinical decision support should consider integrating few-shot techniques to enhance transparency and reduce error rates without additional fine-tuning. \textbf{\textit{Note:}} We did not include \textit{harmlessness} in this evaluation, as the prompts were factual classification queries with minimal risk of eliciting harmful content, focusing instead on correctness and informativeness.

\subsection{Human vs. LLM Judgment Correlation}

Reliable evaluation of clinical AI systems depends on strong alignment between automated metrics and expert human judgment. To assess this, we sampled 200 responses per model and had certified medical annotators independently evaluate them for \textit{honesty}, \textit{helpfulness}, and \textit{harmlessness}. Table~\ref{tab:human_llm_corr} reports the agreement using Cohen's Kappa and Pearson correlation between human ratings and automatic metric predictions. Results show substantial agreement across all three dimensions, with Kappa scores ranging from 0.65 to 0.78 and Pearson correlations between 0.68 and 0.81. \textit{Honesty} showed the strongest alignment, suggesting that automated metrics reliably capture factual alignment. \textit{Helpfulness} and \textit{harmlessness} correlations were slightly lower, indicating room for improving how well current metrics reflect nuanced human judgment in these areas. These findings affirm the value of human-in-the-loop evaluation and support the complementary role of automated tools in scaling clinical AI validation.

\begin{table}[t!]
\centering
\scriptsize
\begin{tabular}{lcc}
\toprule
\textbf{Metric} & \textbf{Kappa} & \textbf{Pearson} \\
\midrule
Honesty & 0.78 & 0.81 \\
Helpfulness & 0.70 & 0.75 \\
Harmlessness & 0.65 & 0.68 \\
\bottomrule
\end{tabular}
\caption{Correlation Between Human Annotator Labels and Automated Metrics: Agreement is highest for honesty, while helpfulness and harmlessness show moderate alignment, highlighting refinement opportunities.}
\label{tab:human_llm_corr}
\end{table}

\begin{figure*}[!t]
\vspace{-0.3cm}
    \centering
    % Row 1
    \begin{subfigure}[b]{0.32\textwidth}
        \centering
        \includegraphics[width=\textwidth]{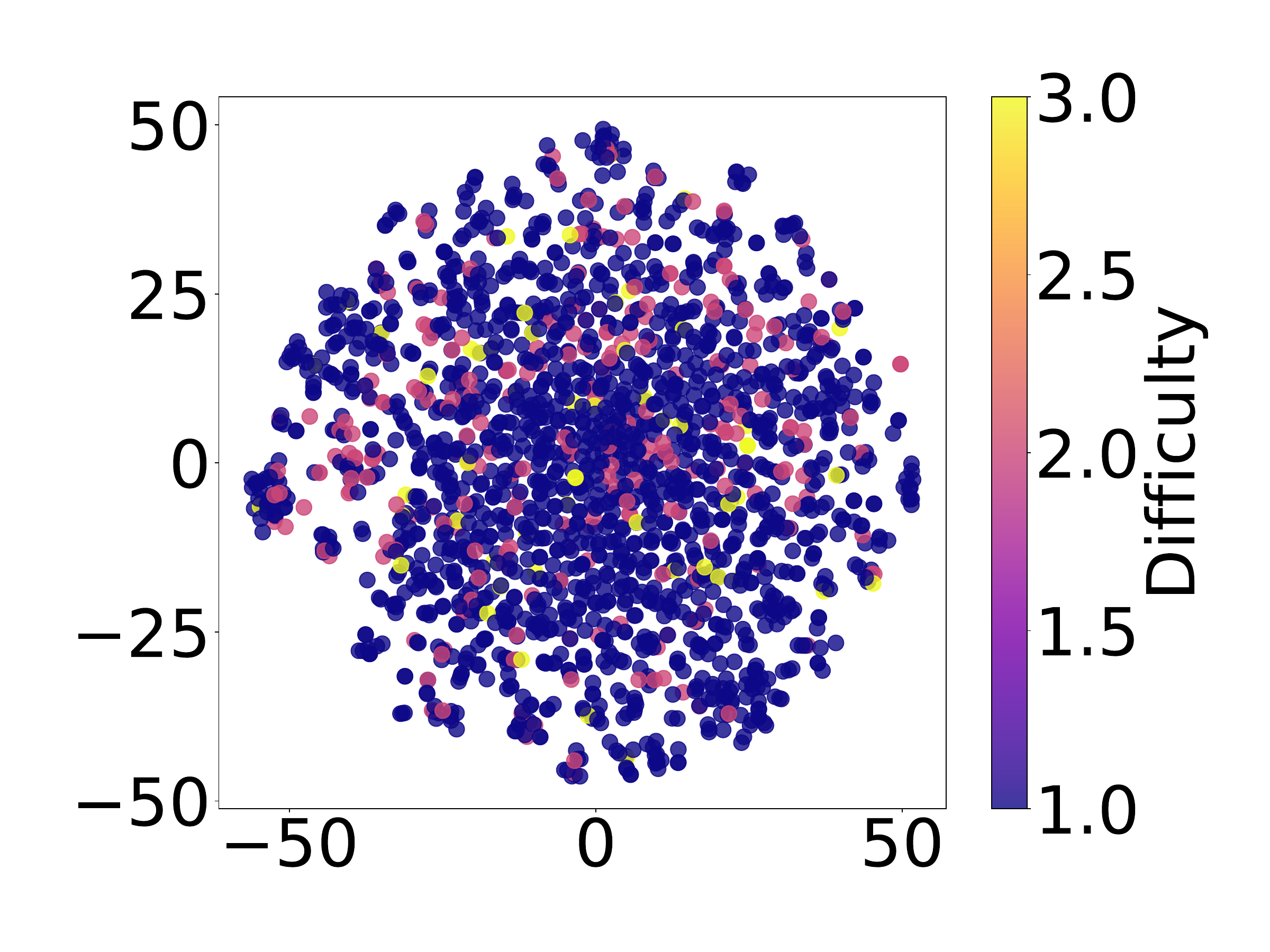}
        \caption{Semantic Distribution via t‑SNE}
    \end{subfigure}
    \hfill
    \begin{subfigure}[b]{0.32\textwidth}
        \centering
        \includegraphics[width=\textwidth]{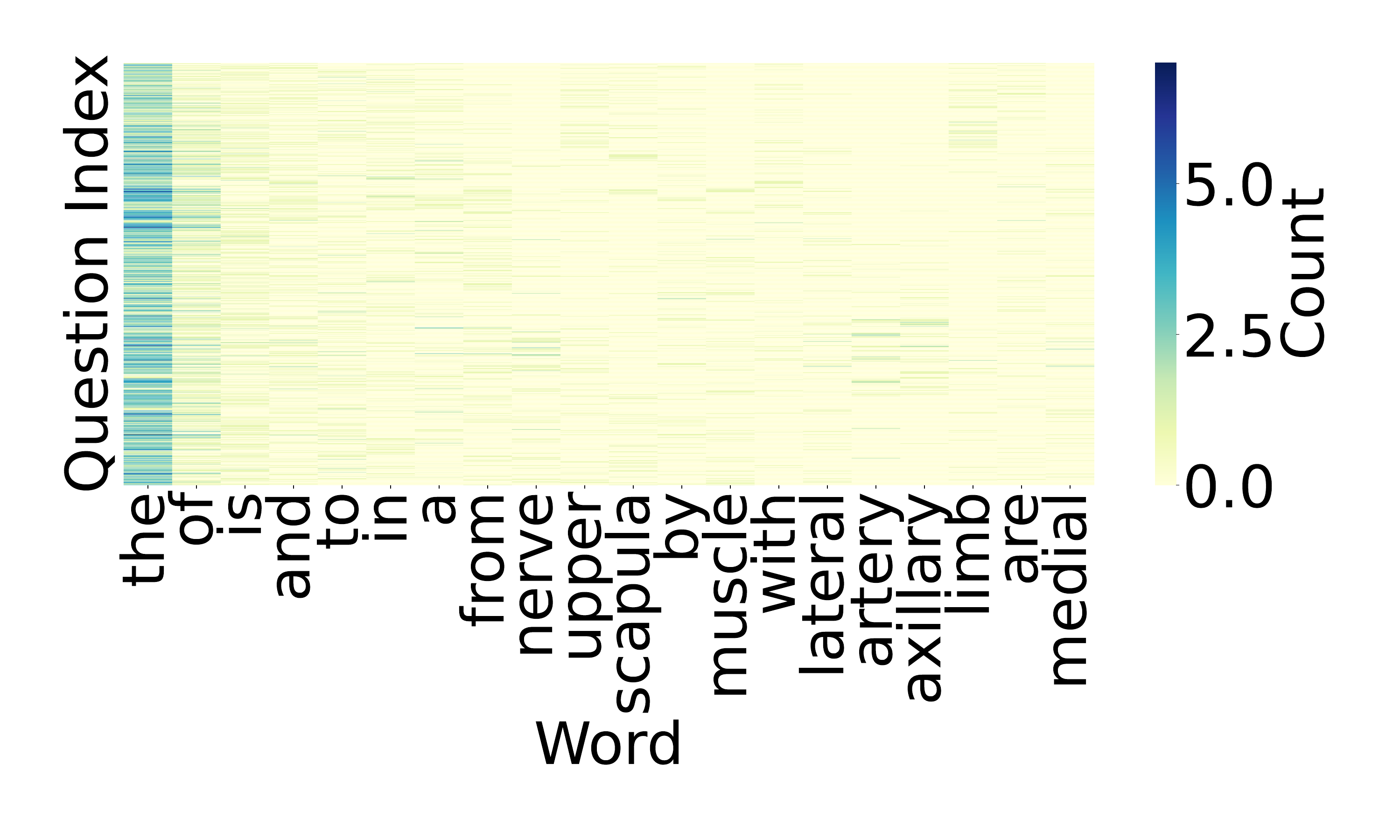}
        \caption{Lexical Heatmap of Question Vocabulary}
    \end{subfigure}
    \hfill
    \begin{subfigure}[b]{0.32\textwidth}
        \centering
        \includegraphics[width=\textwidth]{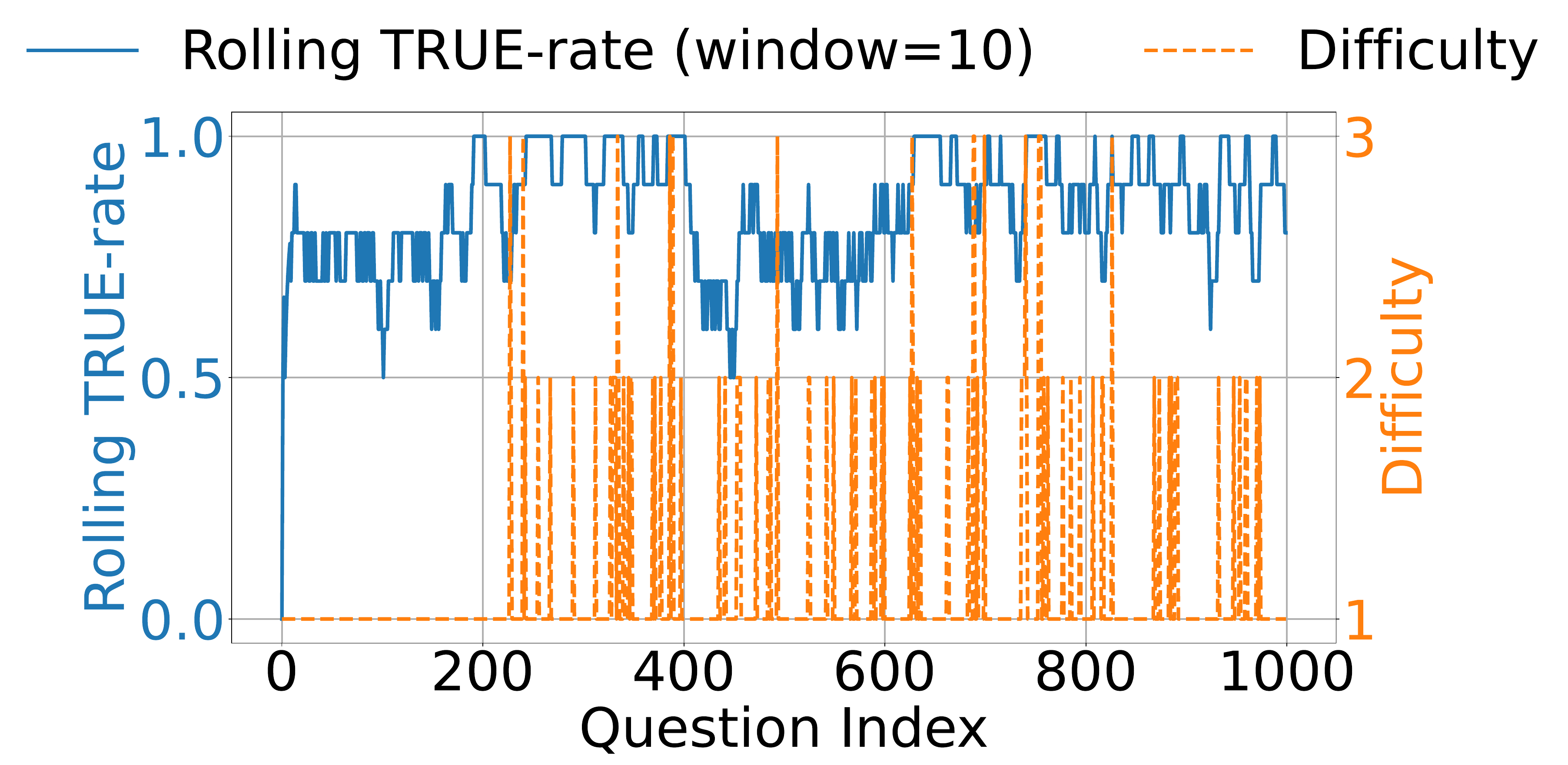}
        \caption{Temporal Dynamics of Correctness and Difficulty}
    \end{subfigure}
    % Row 2
    \begin{subfigure}[b]{0.32\textwidth}
        \centering
        \includegraphics[width=\textwidth]{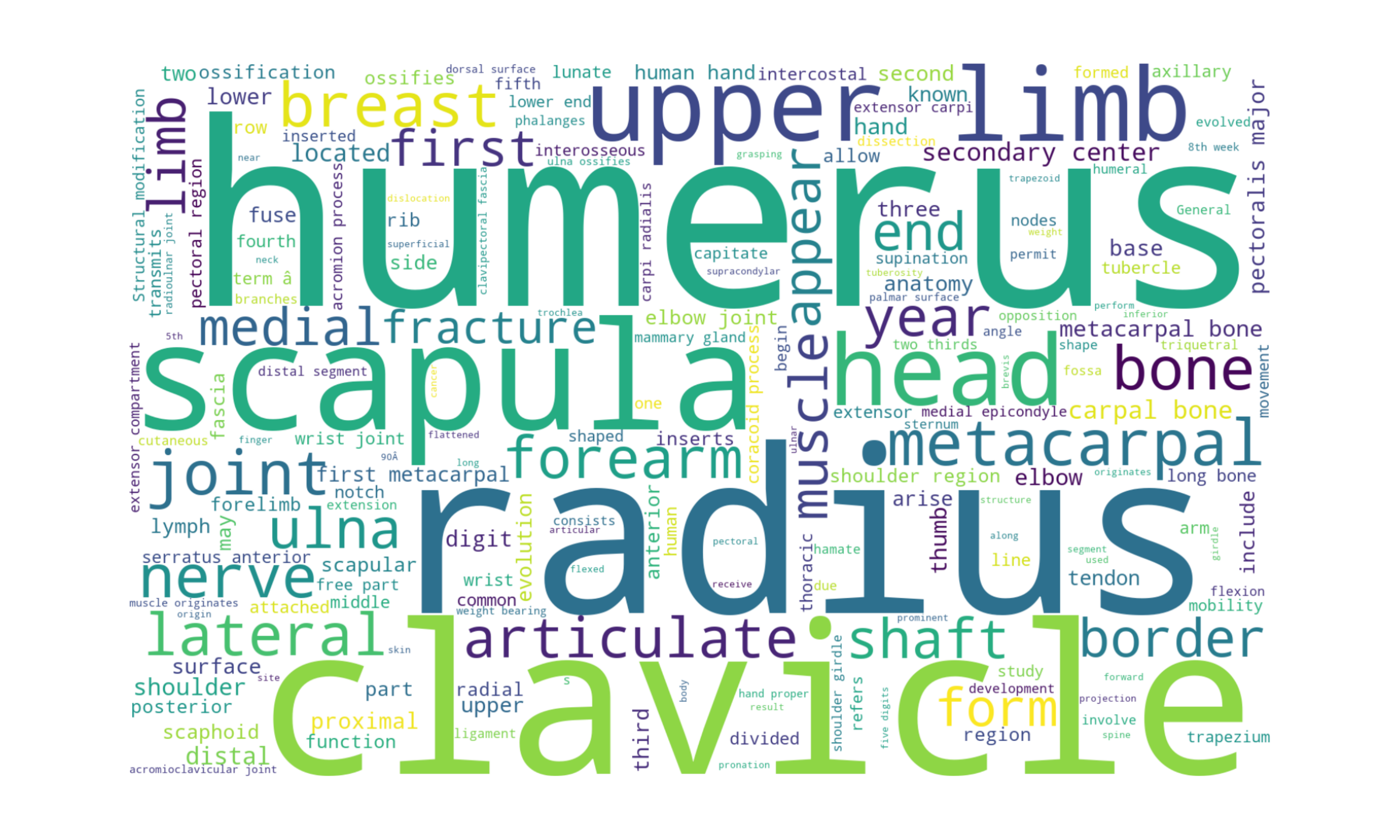}
        \caption{TRUE Answer Word Cloud}
    \end{subfigure}
    \hfill
    \begin{subfigure}[b]{0.32\textwidth}
        \centering
        \includegraphics[width=\textwidth]{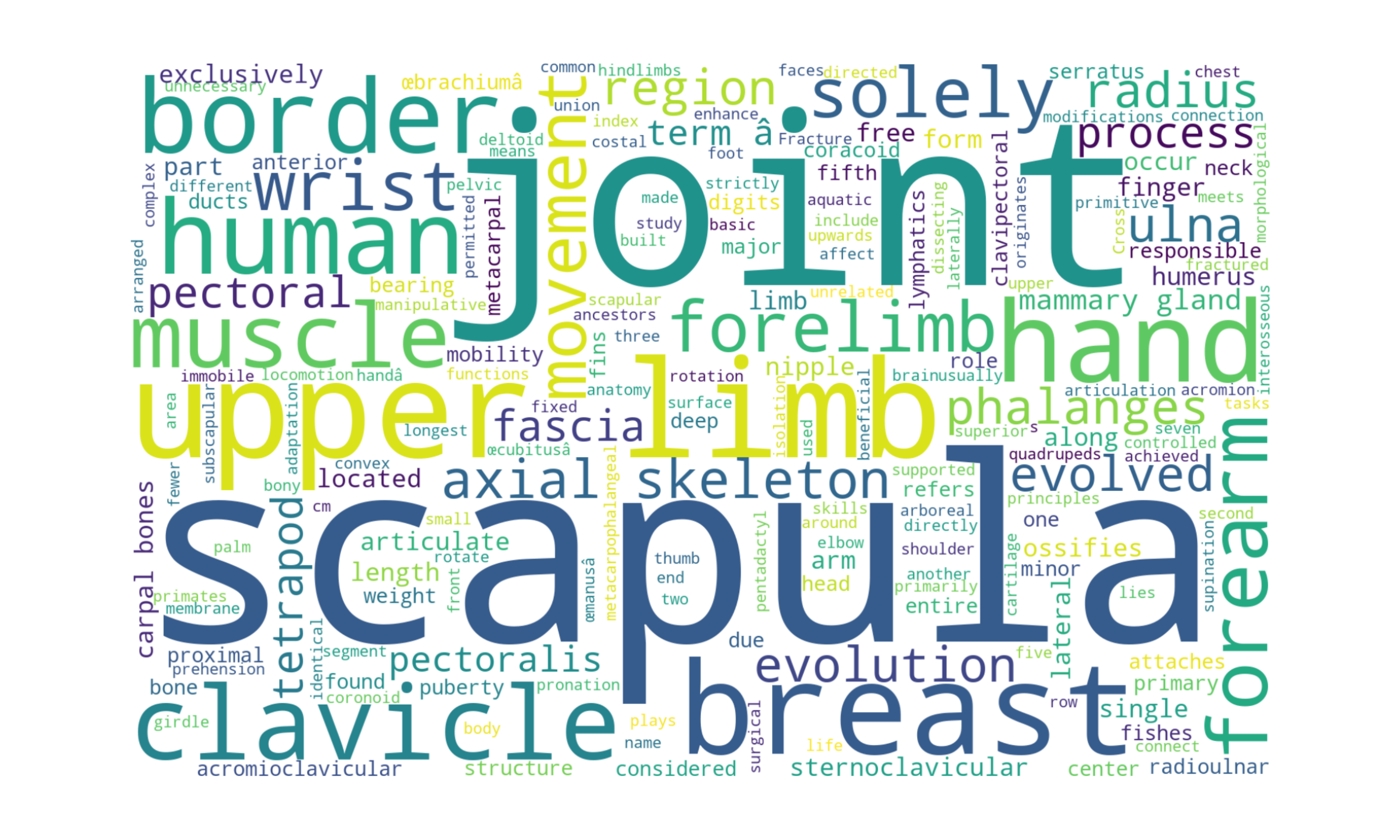}
        \caption{FALSE Answer Word Cloud}
    \end{subfigure}
    \hfill
    \begin{subfigure}[b]{0.32\textwidth}
        \centering
        \includegraphics[width=\textwidth]{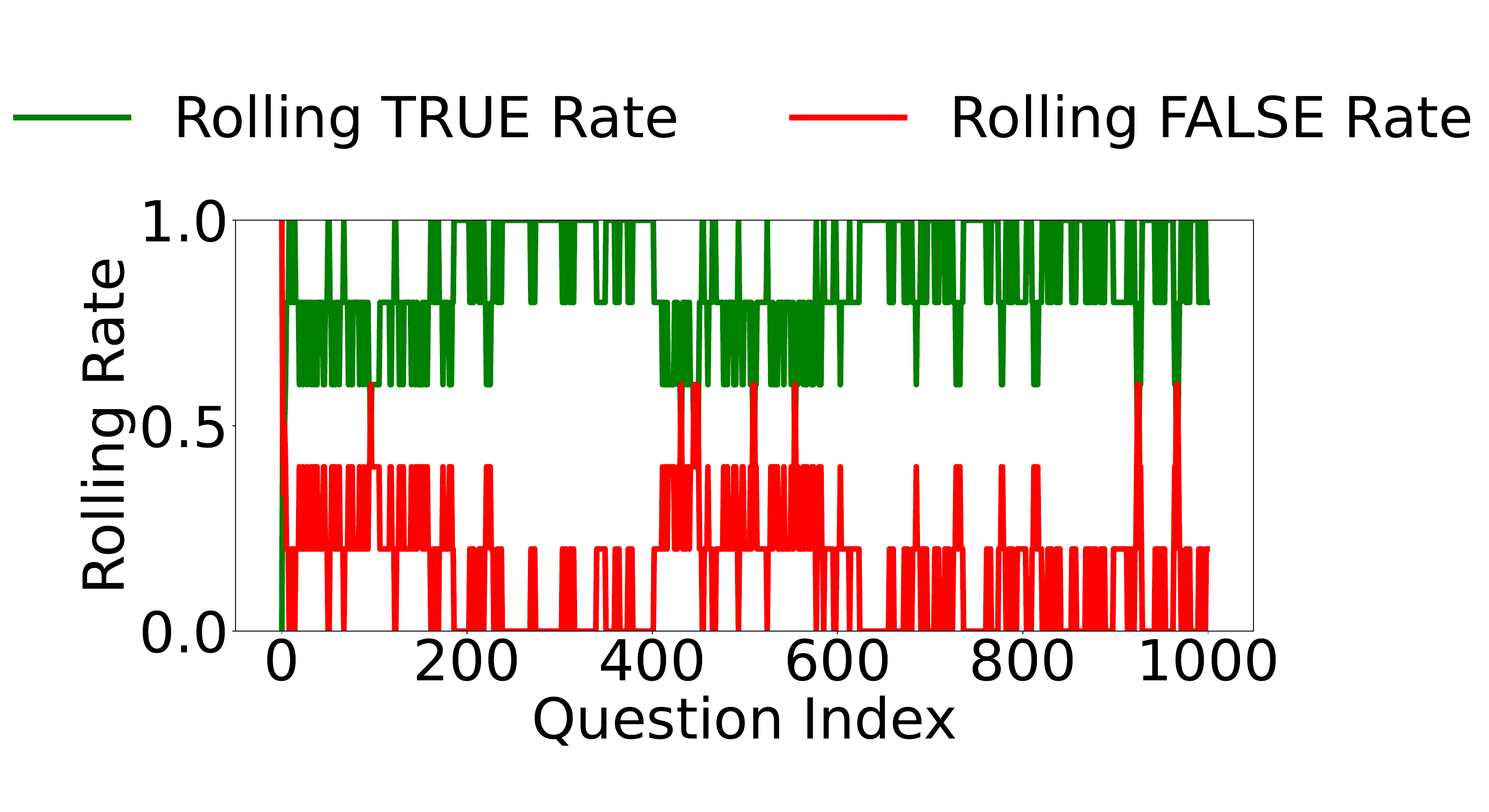}
        \caption{Answer Type Evolution Over Time}
    \end{subfigure}
     % Row 3
    \caption{Semantic and temporal analysis of question-answer behavior. (a) t-SNE shows semantic clustering with difficulty overlay. (b) Heatmap illustrates lexical distribution across question indices. (c) Rolling correctness vs. difficulty trends. (d–e) Word clouds highlight frequent terms in TRUE and FALSE answers. (f) Evolution of answer types over time.}
    \label{Figure 3}
    \vspace{-0.3cm}
\end{figure*}

\paragraph{Qualitative Analysis.} Figure~\ref{Figure 3} offers an in-depth qualitative assessment of the dataset’s semantic and behavioral characteristics. Subfigure (a) uses t-SNE to project the semantic space of question embeddings, revealing dense clustering patterns correlated with difficulty levels—indicating that harder questions tend to occupy semantically distinct regions. The lexical heatmap in (b) highlights word frequency across the question index, showing that specific anatomical terms dominate and vary with question position. Subfigure (c) illustrates temporal dynamics by plotting rolling correctness rates alongside difficulty, uncovering periodic dips in performance that align with more complex or ambiguous question segments. Word clouds in (d) and (e) differentiate lexical emphasis in TRUE and FALSE answers, with TRUE responses focusing on terms like ``radius'', ``clavicle'', and ``scapula'', while FALSE answers include distractors such as ``joint'', ``breast'', and ``border''. Finally, (f) tracks the evolution of answer types over time, showing non-random fluctuations between TRUE and FALSE labels—suggesting shifts in dataset reasoning demands or structural design. Collectively, these visualizations provide insights into the semantic structure, linguistic patterns, and temporal answer behaviors that shape model performance.

\section{Conclusion}

This study evaluated three clinical LLMs---Mistral-7B, BioMistral-7B-DARE, and AlpaCare-13B---on factual accuracy, safety, and reasoning. AlpaCare-13B achieved the best performance with an accuracy of 91.7\% and a \textit{harmlessness} score of 0.92, showcasing its effectiveness in clinical QA. BioMistral-7B-DARE, despite its smaller scale, attained a high safety score of 0.90, highlighting the benefits of domain-specific tuning. Few-shot prompting boosted accuracy from 78\% to 85\%. However, all models exhibited limitations on complex reasoning tasks. These results emphasize persistent challenges in clinical LLMs and the necessity of balancing accuracy, safety, and reasoning for real-world deployment.

\section*{Limitations}
\label{sec:Limitations}

Despite promising results, this study has several limitations. First, the evaluation was restricted to a limited set of clinical LLMs and benchmark datasets, which may not represent the full spectrum of clinical scenarios or model architectures. The reasoning tasks employed were relatively simple, and more complex, real-world clinical reasoning might reveal different performance patterns. Additionally, safety assessments were based on automated metrics and limited human review, which might not capture all nuances of harmful or biased outputs. The study also focused mainly on accuracy, safety, and reasoning but did not evaluate other important aspects such as model interpretability, latency, or resource efficiency, which are critical for clinical deployment. Finally, the few-shot prompting approach improved accuracy but may not generalize across diverse clinical domains or patient populations. Future work should address these limitations by expanding datasets, incorporating more rigorous safety evaluations, and exploring broader clinical applicability.

\section*{Ethics Statement}
\label{sec:Ethics Statement}

This study prioritizes ethical considerations in deploying LLMs in clinical settings. While LLMs hold significant potential to assist healthcare professionals, improper use may lead to misinformation or harm due to incorrect or biased outputs. We emphasize that these models are not substitutes for professional medical advice but tools to augment clinical decision-making. Human oversight remains essential to ensure patient safety and privacy. All evaluated models were tested with anonymized, publicly available clinical questions to avoid exposing sensitive patient information. Moreover, we highlight the need for ongoing monitoring of model behavior to detect and mitigate harmful biases or hallucinations. Our study advocates transparency in reporting model limitations and stresses responsible use to safeguard patient welfare and uphold medical ethics in AI deployment.

\bibliography{main}

\begin{thebibliography}{22}
\providecommand{\natexlab}[1]{#1}

\bibitem[{Benzon(2025)}]{benzon2025llm}
William~L Benzon. 2025.
\newblock From llm mechanisms to ring-composition: A conversation with claude 3.5 working paper.
\newblock \emph{Available at SSRN 5125800}.

\bibitem[{Chang(2023)}]{chang2023examining}
Edward~Y Chang. 2023.
\newblock Examining gpt-4: Capabilities, implications and future directions.
\newblock In \emph{The 10th international conference on computational science and computational intelligence}, pages 1--8.

\bibitem[{Gupta et~al.(2025)Gupta, Singh, Manikandan, and Ehtesham}]{gupta2025digital}
Gaurav~Kumar Gupta, Aditi Singh, Sijo~Valayakkad Manikandan, and Abul Ehtesham. 2025.
\newblock Digital diagnostics: The potential of large language models in recognizing symptoms of common illnesses.
\newblock \emph{Ai}, 6(1):13.

\bibitem[{Jin et~al.(2019)Jin, Dhingra, Liu, Cohen, and Lu}]{jin2019pubmedqa}
Qiao Jin, Bhuwan Dhingra, Zhengping Liu, William Cohen, and Xinghua Lu. 2019.
\newblock Pubmedqa: A dataset for biomedical research question answering.
\newblock In \emph{Proceedings of the 2019 conference on empirical methods in natural language processing and the 9th international joint conference on natural language processing (EMNLP-IJCNLP)}, pages 2567--2577.

\bibitem[{Katz et~al.(2024)Katz, Bommarito, Gao, and Arredondo}]{kim2023does}
Daniel~Martin Katz, Michael~James Bommarito, Shang Gao, and Pablo Arredondo. 2024.
\newblock Gpt-4 passes the bar exam.
\newblock \emph{Philosophical Transactions of the Royal Society A: Mathematical, Physical and Engineering Sciences}, 382(2270).

\bibitem[{Kr{\"a}nzle(2024)}]{kranzle2024evaluating}
Theresa~S Kr{\"a}nzle. 2024.
\newblock \emph{Evaluating Creativity with AI: Comparing GPT Models and Human Experts in Idea Evaluation}.
\newblock Ph.D. thesis, Ph. D. thesis, Copenhagen Business School.

\bibitem[{Kumar et~al.(2024)Kumar, Datta, Singh, Datta, Singh, and Sharma}]{manes2023evaluating}
Sushant Kumar, Sumit Datta, Vishakha Singh, Deepanwita Datta, Sanjay~Kumar Singh, and Ritesh Sharma. 2024.
\newblock Applications, challenges, and future directions of human-in-the-loop learning.
\newblock \emph{IEEE Access}, 12:75735--75760.

\bibitem[{Labrak et~al.(2024)Labrak, Bazoge, Morin, Gourraud, Rouvier, and Dufour}]{labrak2024biomistral}
Yanis Labrak, Adrien Bazoge, Emmanuel Morin, Pierre-Antoine Gourraud, Mickael Rouvier, and Richard Dufour. 2024.
\newblock Biomistral: A collection of open-source pretrained large language models for medical domains.
\newblock In \emph{Findings of the association for computational linguistics: acl 2024}, pages 5848--5864.

\bibitem[{Li et~al.(2023)Li, Meng, Shi, Zhai, and Ruan}]{li2023meddm}
Binbin Li, Tianxin Meng, Xiaoming Shi, Jie Zhai, and Tong Ruan. 2023.
\newblock Meddm: Llm-executable clinical guidance tree for clinical decision-making.
\newblock \emph{arXiv preprint arXiv:2312.02441}.

\bibitem[{Lin et~al.(2022)Lin, Hilton, and Evans}]{lin2022truthfulqa}
Stephanie Lin, Jacob Hilton, and Owain Evans. 2022.
\newblock Truthfulqa: Measuring how models mimic human falsehoods.
\newblock In \emph{Proceedings of the 60th annual meeting of the association for computational linguistics (volume 1: long papers)}, pages 3214--3252.

\bibitem[{L{\'o}pez et~al.(2024)L{\'o}pez, Monoya, and Espinosa}]{lopez2024design}
IM~Garc{\'\i}a L{\'o}pez, MS~Ram{\'\i}rez Monoya, and JM~Molina Espinosa. 2024.
\newblock Design and challenges of open large language model frameworks (open llm): A systematic literature mapping.
\newblock \emph{ICERI2024 Proceedings}, pages 10320--10328.

\bibitem[{Pal et~al.(2022)Pal, Umapathi, and Sankarasubbu}]{pal2022medmcqa}
Ankit Pal, Logesh~Kumar Umapathi, and Malaikannan Sankarasubbu. 2022.
\newblock Medmcqa: A large-scale multi-subject multi-choice dataset for medical domain question answering.
\newblock In \emph{Conference on health, inference, and learning}, pages 248--260. PMLR.

\bibitem[{Samo et~al.(2024)Samo, Ali, Memon, Abbasi, Koondhar, and Dahri}]{samo2024fine}
Hassan Samo, Kashif Ali, Muniba Memon, Faheem~Ahmed Abbasi, Muhammad~Yaqoob Koondhar, and Kamran Dahri. 2024.
\newblock Fine-tuning mistral 7b large language model for python query response and code generation: A parameter efficient approach.
\newblock \emph{VAWKUM Transactions on Computer Sciences}, 12(1):205--217.

\bibitem[{Singh(2013)}]{singh2024selective}
Vishram Singh. 2013.
\newblock \emph{Selective Anatomy: Prep Manual for Undergraduates}.
\newblock Elsevier India.

\bibitem[{Singhal et~al.(2023)Singhal, Azizi, Tu, Mahdavi, Wei, Chung, Scales, Tanwani, Cole-Lewis, Pfohl et~al.}]{singhal2023large}
Karan Singhal, Shekoofeh Azizi, Tao Tu, S~Sara Mahdavi, Jason Wei, Hyung~Won Chung, Nathan Scales, Ajay Tanwani, Heather Cole-Lewis, Stephen Pfohl, et~al. 2023.
\newblock Large language models encode clinical knowledge.
\newblock \emph{Nature}, 620(7972):172--180.

\bibitem[{Tu et~al.(2024)Tu, Azizi, Driess, Schaekermann, Amin, Chang, Carroll, Lau, Tanno, Ktena et~al.}]{tu2024towards}
Tao Tu, Shekoofeh Azizi, Danny Driess, Mike Schaekermann, Mohamed Amin, Pi-Chuan Chang, Andrew Carroll, Charles Lau, Ryutaro Tanno, Ira Ktena, et~al. 2024.
\newblock Towards generalist biomedical ai.
\newblock \emph{Nejm Ai}, 1(3):AIoa2300138.

\bibitem[{Vaishya(2024)}]{vaishya2024dr}
Raju Vaishya. 2024.
\newblock Dr bhagwan din chaurasia: A guiding light and a pillar of anatomy education in india.
\newblock \emph{Apollo Medicine}, 21(4):381--385.

\bibitem[{Wang et~al.(2025)Wang, Huang, and Chen}]{wang2025mesaqa}
Jui-I Wang, Hen-Hsen Huang, and Hsin-Hsi Chen. 2025.
\newblock Mesaqa: A dataset for multi-span contextual and evidence-grounded question answering.
\newblock In \emph{Proceedings of the 31st International Conference on Computational Linguistics}, pages 10891--10901.

\bibitem[{Yang et~al.(2025)Yang, Chen, Guo, Chen, Lin, Hu, Hu, Wu, and Wang}]{yang2024llm}
Hang Yang, Hao Chen, Hui Guo, Yineng Chen, Ching-Sheng Lin, Shu Hu, Jinrong Hu, Xi~Wu, and Xin Wang. 2025.
\newblock Llm-medqa: Enhancing medical question answering through case studies in large language models.
\newblock In \emph{2025 International Joint Conference on Neural Networks (IJCNN)}, pages 1--8. IEEE.

\bibitem[{Zhang et~al.(2026)Zhang, Jiang, Chi, Chen, Elkoumy, Wang, Wu, Zhou, Pan, Wang et~al.}]{zhang2025diagnosing}
Liangliang Zhang, Zhuorui Jiang, Hongliang Chi, Haoyang Chen, Mohammed Elkoumy, Fali Wang, Qiong Wu, Zhengyi Zhou, Shirui Pan, Suhang Wang, et~al. 2026.
\newblock Diagnosing and addressing pitfalls in kg-rag datasets: Toward more reliable benchmarking.
\newblock \emph{Advances in Neural Information Processing Systems}, 38.

\bibitem[{Zhang et~al.(2023)Zhang, Tian, Yang, Chen, Li, and Petzold}]{zhang2023alpacare}
Xinlu Zhang, Chenxin Tian, Xianjun Yang, Lichang Chen, Zekun Li, and Linda~Ruth Petzold. 2023.
\newblock Alpacare: Instruction-tuned large language models for medical application.
\newblock \emph{arXiv preprint arXiv:2310.14558}.

\bibitem[{Zheng et~al.(2023)Zheng, Chiang, Sheng, Zhuang, Wu, Zhuang, Lin, Li, Li, Xing et~al.}]{zheng2023judging}
Lianmin Zheng, Wei-Lin Chiang, Ying Sheng, Siyuan Zhuang, Zhanghao Wu, Yonghao Zhuang, Zi~Lin, Zhuohan Li, Dacheng Li, Eric Xing, et~al. 2023.
\newblock Judging llm-as-a-judge with mt-bench and chatbot arena.
\newblock \emph{Advances in neural information processing systems}, 36:46595--46623.

\end{thebibliography}

\end{document}